\documentclass[a4paper]{article}

\usepackage{INTERSPEECH_v2}

\usepackage{graphicx}
\usepackage{amssymb,amsmath,bm}

\usepackage{times}
\ninept

\title{Comparing Human and Machine Errors in Conversational Speech Transcription}

\name{Andreas Stolcke \quad Jasha Droppo}
\address{
  Microsoft AI and Research, Redmond, WA, USA}
\email{anstolck@microsoft.com, jdroppo@microsoft.com}

\begin{document}

\maketitle
\begin{abstract}
  Recent work in automatic recognition of conversational telephone speech (CTS) has achieved accuracy levels comparable to human
  transcribers, although there is some debate how to precisely quantify human performance on this task, using
  the NIST 2000 CTS evaluation set.
  This raises the question what systematic differences, if any, may be found differentiating human from machine transcription errors.
  In this paper we approach this question by comparing the output of our most accurate CTS recognition system to that of a standard
  speech transcription vendor pipeline.
  We find that the most frequent substitution, deletion and insertion error types of both outputs show a high degree of overlap.
  The only notable exception is that the automatic recognizer tends to confuse filled pauses (``uh'')
  and backchannel acknowledgments (``uhhuh'').  Humans tend not to make this error, presumably due to the distinctive and opposing 
  pragmatic functions attached to these words.
  Furthermore, we quantify the correlation between human and machine errors at the speaker level, 
  and investigate the effect of speaker overlap between training and test data.
  Finally, we report on an informal ``Turing test'' asking humans to discriminate between automatic and human transcription error
  cases.
\end{abstract}


\section{Introduction}

Automatic speech recognition (ASR) systems have seen remarkable advances over the last half-decade from the use of 
deep, convolutional and recurrent neural network architectures, enabled by a combination of modeling advances,
available training data, and increased computational resources.
Given these advances, our research group recently embarked on an effort to reach human-level transcription
accuracy using state-of-the-art ASR techniques on one of the genres of speech that has historically served as a
difficult benchmark task: conversational telephone speech (CTS).
About a decade ago, CTS recognition had served as an evaluation task for government-sponsored work in speech recognition,
predating the take-over of deep learning approaches and still largely in the GMM-HMM modeling framework
\cite{chen2006advances,matsoukas2006advances,stolcke2006recent,ljolje2001,gauvain2003conversational,evermann2004development}.
It had proven to be a hard problem, due to the variable nature of conversational pronunciations, speaking styles,
and regional accents.
Seide at al.~\cite{seide2011conversational} demonstrated that deep networks as acoustic models could achieve significant
improvements over GMM-HMM models on CTS data, and more recently researchers at IBM had achieved results on this task
that represented a further significant advance \cite{saon2015ibm,saonSRK16} over those from a decade ago.

The goal of reaching ``human parity'' in automatic CTS transcription raises the question of what should be considered human
accuracy on this task.
We operationalized the question by submitting the chosen test data to the same vendor-based transcription pipeline that 
is used at Microsoft for production data (for model training and internal evaluation purposes), and then comparing the results
to ASR system output under the NIST scoring protocol.
Using this methodology, and incorporating state-of-the-art convolutional and recurrent network architectures for both
acoustic modeling \cite{sak2014long,sak2015fast,saon2015ibm,sercu2016very,bi2015very,qian2016very,yu2016deep} 
and language modeling \cite{mikolov2010recurrent,mikolov2012context,sundermeyer2012lstm,medennikov2016improving} with 
extensive use of model combination, we obtained a machine error rate that was very slightly below that of the human transcription
process (5.8\% versus 5.9\% on Switchboard data, and 11.0\% versus 11.3\% on CallHome English data) \cite{parity-techreport}.
Since then, Saon et al.\ have reported even better results, along with a separate transcription experiment that puts the 
human error rate, on the same test data, at a lower point than measured by us (5.1\% for Switchboard, 6.8\% for CallHome)
\cite{ibm-human:arxiv2017}.

In this paper, we address the question whether there are major qualitative differences between the results of human
transcriptions of conversational speech and those obtained by ASR systems, based on a detailed analysis of the data and system
output from our human parity experiment \cite{parity-techreport}.
The question becomes important if ASR is to replace humans as the first step in fully automatic speech understanding systems%
---if machine transcription errors are qualitatively different from humans then we would have to worry about
the possible effects on downstream processing, and mitigation techniques so as to still achieve an overall ``natural''
user experience (e.g., in real-time conversational speech translation, such as in the Skype application).

We start by discussing why human error rate on this task must themselves be considered a moving target.
Next we ask whether speech that is difficult for ASR also tends to be hard for humans to transcribe (and vice-versa),
and whether the speaker overlap with the training data that is found in a portion of the test data has a noticeable effect 
on the result, as was suggested in \cite{ibm-human:arxiv2017}.
We then look at the most frequent word error types exhibited by the two transcription systems (human and machine),
and finally report on a very preliminary but still informative experiment to see if humans could tell apart the transcription
source (again, human versus machine), based on the errors they make.

\begin{figure*}[t]
\centering
\includegraphics[width=0.45\textwidth]{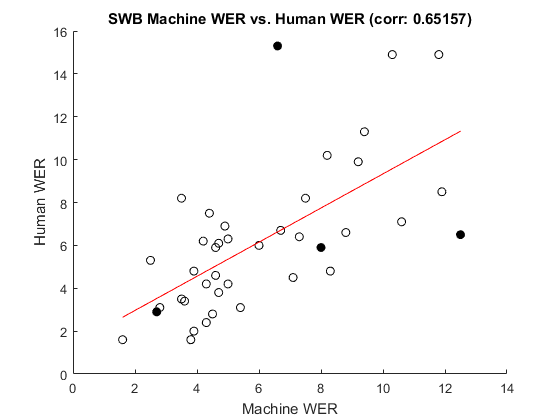}
\quad%
\includegraphics[width=0.45\textwidth]{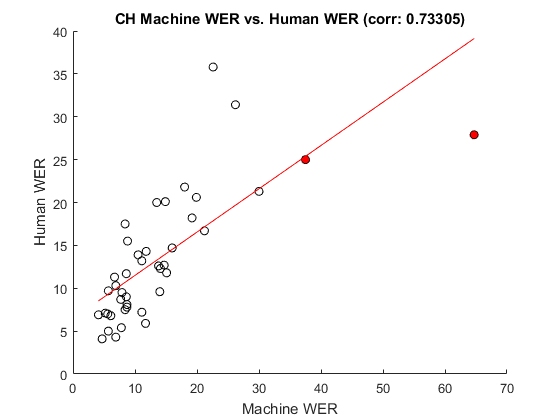}
\caption{Correlation between machine and human word error rates at speaker level.
The solid black circles represent SWB speakers {\em not} seen in training.
The solid red circles stand for secondary CH speakers that share a conversation side with a dominating primary speaker.}
\label{fig:wer-corr}
\end{figure*}

\section{Measuring Human Error}

The assessment of human transcription error on conversational speech has been somewhat murky.
A widely cited figure is 4\% word error rate (WER), based on \cite{lippmann1997speech}.
However, the reference therein is only a ``personal communication'' without further data.
The Linguistics Data Consortium quantified inter-transcriber disagreement for the NIST 2003 CTS evaluation data 
at between 4.1\% and 4.5\% with very careful multiple transcriptions \cite{glenn2010transcription}.
For ``quick transcription'', the disagreement increased to 9.6\%.
The CTS data in the NIST study is from the Switchboard (SWB) and Fisher corpora, and is therefore comparable to the SWB portion
of our data, i.e., coming from telephone conversations between strangers discussing a general-interest topic.
Still, the exact dataset is different, which may account for some of the discrepancy with error rates measured on the
NIST 2000 set used by us (5.9\%) and IBM (5.1\%), although the numbers are remarkably close.

As briefly described in the introduction, we measured human performance by leveraging an existing pipeline in which 
Microsoft data is transcribed on a weekly basis. This pipeline uses 
a large commercial vendor to perform two-pass transcription. In the first pass,
a transcriber works from scratch to transcribe the data. In the second pass,
a second listener monitors the data to do error correction. Dozens of hours
of test data are processed in each batch, with no special instructions to the transcribers.
The waveform segments, roughly corresponding to utterances, making up the test set are processed separately.
This makes the task easier since the speakers are more clearly separated, but also more
difficult since the two sides of the conversation are not interleaved and context may be missing.%
\footnote{This happens to be the way our ASR system operates since the language model currently has no context across utterances.
Still, we realize this deprives humans of a potential advantage over current technology.}
We performed that text normalization on the human transcripts to remove systematic discrepancies with the NIST
scoring references.
(Since this was done with some amount of trial and error it effectively was ``cheating''
for the benefit of the human transcribers.)
We then applied the NIST scoring tools to obtain word error rates of 5.9\% on the SWB portion, and 
11.3\% on the CallHome (CH) portion of the NIST 2000 test set.  
The latter corpus, unlike Switchboard, consists of conversations between friends and family, without seed topic, which 
would account for the much higher overall error rate.
Clearly our method was not designed to achieve the highest possible human transcription accuracy;  instead, as 
pointed out in \cite{parity-techreport}, our goal was to establish a benchmark corresponding to industry-standard
(i.e. high-volume) professional transcript production.

The authors in \cite{ibm-human:arxiv2017} undertook to measure human error on the same dataset, but using a more involved process.
The major differences were:
(1) The transcription vendor was cognizant of the experiment and actively involved. (2) Transcribers were chosen based on past
performance and familiarized with the conventions used by LDC in generating the reference transcripts.
(3) Three independent, parallel transcribers were used, plus a fourth one for 2nd-pass quality control (QC) of the 1st-pass output.
All in all, the transcribers performed roughly 12 to 18 listening passes.
(4) The final output was obtained by choosing the transcriber (with QC) who had obtained the lowest WER on the test data.
As noted earlier, the resulting WERs were 5.1\% and 6.8\%, respectively.   
The considerably lower estimate for CH could be a result of the transcribers having access to the entire conversation (as per personal communication with the authors).  This would be especially helpful in transcribing unfamiliar vocabulary and speaking
styles (allowing the transcriber to ``adapt'' to the data more effectively).

Clearly the IBM experiment made a much more thorough effort to probe the boundaries of human accuracy,
and may in fact have come close to the inter-transcriber agreement previously measured by LDC on a different data set.
However, it is important to realize that further improvements on the human side are no doubt achievable.
For example, the number of transcribers could be scaled up further, or they could be allowed to confer with each other,
to resolve disagreements.  This raises the question of where to draw the line on human effort.

Finally, it is important to realize that conversational speech has a high degree of inherent ambiguity.
For example, conversational pronunciations are highly variable and often reduced \cite{GreenbergEtAl:icslp96}.
Another source of ambiguity is the lack of context and knowledge shared by the speakers (especially in the case of CH).
In the presence of inherent ambiguity, inter-transcriber agreement can be improved by agreed-upon disambiguation rules,
although this would not necessarily reflect true agreement based on speech understanding.

\section{Machine Transcription System}

The details of our conversational speech recognition system are described elsewhere \cite{parity-techreport},
so we only give a brief summary here.
The system employs independent decodings by diverse acoustic models, including convolutional neural net (CNN) 
and bidirectional long short-term memory (BLSTM) models that differ by model architecture,
number of senones, amount of training data, and other metaparameters.
Decoding uses a pruned 4-gram N-gram language model (LM) to generate lattices, which are then expanded into
500-best lists using a larger N-gram LM.
The N-best lists are rescored with multiple LSTM-LMs operating in forward and backward directions. 
Model scores are combined log-linearly at the utterance level and converted to posterior probabilities
represented as word confusion networks.
The various subsystems making up the final system are selected in a greedy search, and their weights are
optimized via an expectation-maximization algorithm, on development data.
The acoustic training data comprises all the publicly available CTS data (about 2000 hours), while the LMs are
additionally trained on Broadcast News and Web data from U. Washington.
The individual subsystems (based on different acoustic models) achieve word error rates between 6.4\% and 7.7\% on the
Switchboard evaluation set, and between 12.2\% and 17.0\% on the CallHome portion.
Combined, the system achieves 5.8\% and 11.0\% WER, respectively.

\section{Error Distribution and Correlation}

\begin{table*}[t]
    \centering
\caption{Most common substitutions for ASR system and humans. The number of
times each error occurs is followed by the word in the reference, and what
appears in the hypothesis instead.}
\vspace*{0.1in}
\label{tab:subs}
        \small
    \begin{tabular}{|l|l||l|l|}
    \hline
    \multicolumn{2}{|c||}{{\bf CH}} & \multicolumn{2}{c|}{{\bf SWB}} \\ \hline
                    {\bf ASR}   & {\bf Human}   & {\bf ASR}    & {\bf Human}   \\ \hline \hline
45:  (\%hesitation) / \%bcack & 12:  a / the & 29:  (\%hesitation) / \%bcack & 12:  (\%hesitation) / hmm \\ \hline
12:  was / is & 10:  (\%hesitation) / a &  9:  (\%hesitation) / oh & 10:  (\%hesitation) / oh \\ \hline
9:  (\%hesitation) / a & 10:  was / is &  9:  was / is &  9:  was / is \\ \hline
8:  (\%hesitation) / oh & 7:  (\%hesitation) / hmm &  8:  and / in &  8:  (\%hesitation) / a \\ \hline
8:  a / the & 7:  bentsy / bensi &  6:  (\%hesitation) / i &  5:  in / and \\ \hline
7:  and / in & 7:  is / was &  6:  in / and &  4:  (\%hesitation) / \%bcack \\ \hline
7:  it / that & 6:  could / can &  5:  (\%hesitation) / a &  4:  and / in \\ \hline
6:  in / and & 6:  well / oh &  5:  (\%hesitation) / yeah &  4:  is / was \\ \hline
5:  a / to & 5:  (\%hesitation) / \%bcack &  5:  a / the &  4:  that / it \\ \hline
5:  aw / oh & 5:  (\%hesitation) / oh &  5:  jeez / jeeze &  4:  the / a \\ \hline
        \end{tabular}
\end{table*}

\begin{table}[t]
    \centering
\caption{Most common deletions for ASR system and humans.}
\vspace*{0.1in}
\label{tab:dels}
        \small
    \begin{tabular}{|l|l||l|l|}
    \hline
    \multicolumn{2}{|c||}{{\bf CH}} & \multicolumn{2}{c|}{{\bf SWB}} \\ \hline 
                    {\bf ASR}   & {\bf Human}   & {\bf ASR}    & {\bf Human}   \\ \hline \hline
 44:  i  &     73:  i &    31:  it &    34:  i \\ \hline
      33:  it &       59:  and &       26:  i &       30:  and \\ \hline
      29:  a &       48:  it &       19:  a &       29:  it \\ \hline
      29:  and &       47:  is &       17:  that &       22:  a \\ \hline
      25:  is &       45:  the &       15:  you &       22:  that \\ \hline
      19:  he &       41:  \%bcack &       13:  and &       22:  you \\ \hline
      18:  are &       37:  a &       12:  have &       17:  the \\ \hline
      17:  oh &       33:  you &       12:  oh &       17:  to \\ \hline
      17:  that &       31:  oh &       11:  are &       15:  oh \\ \hline
     17:  the &      30:  that &      11:  is &      15:  yeah \\ \hline
        \end{tabular}
\end{table}

\begin{table}[t]
    \centering
\caption{Most common insertions for ASR system and humans.}
\vspace*{0.1in}
\label{tab:ins}
        \small
    \begin{tabular}{|l|l||l|l|}
    \hline
    \multicolumn{2}{|c||}{{\bf CH}} & \multicolumn{2}{c|}{{\bf SWB}} \\ \hline 
                    {\bf ASR}   & {\bf Human}   & {\bf ASR}   & {\bf Human}  \\ \hline \hline
15:  a &    10:  i &    19:  i &    12:  i \\ \hline
      15:  is &        9:  and &        9:  and &       11:  and \\ \hline
      11:  i &        8:  a &        7:  of &        9:  you \\ \hline
      11:  the &        8:  that &        6:  do &        8:  is \\ \hline
      11:  you &        8:  the &        6:  is &        6:  they \\ \hline
       9:  it &        7:  have &        5:  but &        5:  do \\ \hline
       7:  oh &        5:  you &        5:  yeah &        5:  have \\ \hline
       6:  and &        4:  are &        4:  air &        5:  it \\ \hline
       6:  in &        4:  is &        4:  in &        5:  yeah \\ \hline
      6:  know &       4:  they &       4:  you &     4:  a \\ \hline
        \end{tabular}
\end{table}

We note in passing that machine and human transcription WERs do not differ significantly according the Wilcoxon and 
Matched Pairs Sentence Segment Word Error tests as applied by NIST, nor do they differ according to a Sign test comparing error
counts at the utterance level.

A first high-level question regarding the relation between word errors by machine and human transcribers is whether difficulty
in one predicts difficulty in the other.
Figure~\ref{fig:wer-corr} shows scatter plots of speaker-level error rates (machine vs.~human), separated by corpus.
Each corpus subset has 40 conversation sides.

Clearly the errors at that level are correlated, with $\rho = 0.65$ for SWB and $\rho = 0.73$ for CH.
This suggests that properties of the speech, either as a function of the content, the speaker, or the channel
(each speaker occurs in exactly one
test conversation), cause errors for both machine and human transcription.

We observe that the CH data has two speakers with outlier machine error rates
(37.5\% and 64.7\% WER, solid red dots in Figure~\ref{fig:wer-corr}).
These correspond to secondary speakers in their respective conversation sides,
each with only a fraction of the speech of the dominant speaker.
Note that the ASR system processes each conversation assuming only a single speaker per side.  If we remove these outliers, the 
machine-human error correlation on CH increases to $\rho = 0.80$.
With secondary speakers excluded, we can also observe that the machine error rates cluster tighter than the human ones
in both corpora (SWB: machine $6.1\% \pm 2.8$ vs.\ human $6.2\% \pm 3.4$; CH: machine $11.8\% \pm 6.1$ vs.\ human $12.7\% \pm 7.0$).

In \cite{ibm-human:arxiv2017} it was sugggested that one of the reasons for the much higher error rate on CH compared to SWB 
was that 36 of the 40 SWB test speakers occur in the portion of the SWB corpus that is used in training (due to what we
surmise to be an oversight in the selection of the NIST 2000 test set).
To assess this hypothesis we singled out the four speakers in the SWB portion that are not found in the training set; these are shown as solid black circles in Figure~\ref{fig:wer-corr}.
At first, it seems that the speaker-averaged WER for the ``seen'' speakers (machine WER 5.9\%) is indeed much lower than for the speakers not found in training (7.5\%).
However, we can safely attribute this to bad luck and small sample size.  The average machine WER of 7.5\% for ``unseen'' speakers
is well within one standard deviation of the ``seen'' speakers' WER distribution ($5.9\% \pm 2.7$), and more tellingly, 
almost exactly the same relative difference in WERs between ``seen'' and ``unseen'' speakers is observed for human transcriptions
(6.0\% versus 7.7\%).
Clearly the human transcribers did not have the benefit of training on the ``seen'' speakers,
so the difference must be due to the intrinsic difficulty of the speakers, which affects both transcription systems.

\section{Error types}

Tables~\ref{tab:subs}--\ref{tab:ins} show the top ten types of substitutions, deletions and insertions
for both ASR and human transcripts.
Inspections reveals that the same short function words, discourse markers and filled pauses appear in the top ten errors
for both systems.
There is one notable exception, however.
The top substitution error for the ASR system involves misrecognition of filled pauses (``\%hesitation'', a word class label
covering ``uh'' and ``um'' in various spellings) as backchannel acknowledgments (``\%bcack'', standing for ''uhhuh'', ``mhm'',
etc.).\footnote{Note that the reverse substitution cannot occur because we programmatically delete filled pauses
from the recognizer output, since they can only increase recognition errors under the NIST scoring protocol.}
The same substitution error is much less frequent in human transcripts.

A possible explanation for this asymmetry lies in the discourse functions of filled pauses and backchannels.
Filled pauses serve to either claim or retain the floor, signaling that the speaker wants to either start or continue speaking.
Backchannels, on the other hand, acknowledge that the speaker is listening, and that the other speaker should carry on.
Since the two classes of words thus have exactly opposite functions in turn management, it stands to reason that humans are keenly aware of their differences and use all available phonetic, prosodic, and contextual cues to distinguish then.
Our ASR system, by contrast, uses only its standard acoustic-phonetic and language models.
Modeling dialog context in particular would be expected to improve this shortcoming.

\section{A Turing-like Experiment}

\begin{figure}[t]
\centering
\includegraphics[width=0.5\textwidth]{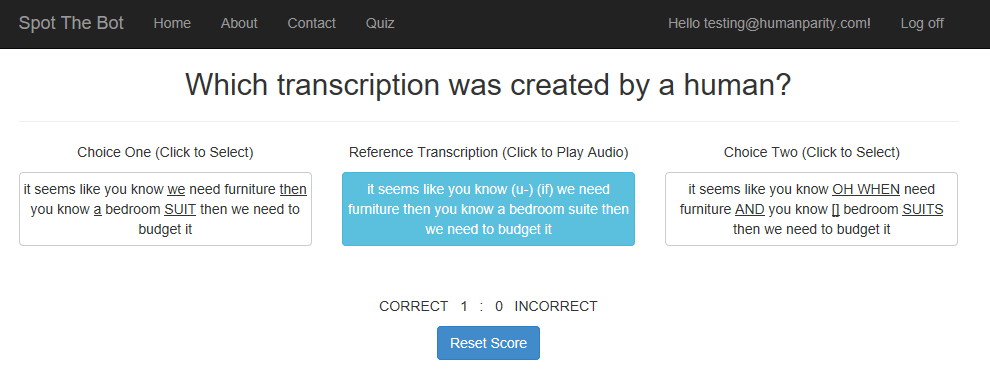}
\caption{Turing-like test challenging human players to tell machine from human transcripts}
\label{fig:spot-the-bot}
\end{figure}

Having established that human and machine transcriptions are quite similar in several aspects, including the word token types 
involved, we were wondering if higher-level error patterns could distinguish the two systems.
For example, one might expect that human misrecognitions are guided by a strong ``human'' language and understanding model, whereas
machine errors might be more likely to generate syntactic and semantic nonsense.
To get at this question we designed a specialized version of the classic Turing test, in the sense that a human judge is asked to 
interact with a system with the goal of estimating whether it is underpinned by human or artificial ``intelligence.''
In our case, the task involved inspecting one randomly chosen utterance from the test set at a time, with a side-by-side
display of the reference transcript, the human transcript, and the ASR output (after the text normalizations that are part of the
scoring protocol).
Only utterances having at least one transcription error and a discrepancy between the two versions are presented.
Discrepancies between the transcript versions are highlighted, and the error type (substitution, insertion, deletion) is visually
coded as well, as shown in Figure~\ref{fig:spot-the-bot}.

We ran this informal experiment during four days on the exhibitor floor of the 2017 IEEE ICASSP conference in New Orleans.%
\footnote{The experimental setup also supports audio playback of the test utterances.  However, this function was not used due to technical difficulties.}
The players were not formally recruited or characterized, but consisted of conference attendees who for the most part had some
background or experience in speech processing.
Subjects were introduced to the test by explaining the research background, and  were allowed to play as many trials
as they wanted.
Out of a total of 353 trials, subjects identified the human transcript correctly 188 times,
for an overall success rate of 53\%.
The successes included occasional gimmes like human misspellings  or the asymmetry in the filled pause/backchannel substitution
(which we pointed out to the subjects).
According to a binomial test, this success rate does not differ signficantly from the 50\% chance rate
($p \approx 0.12$, one-tailed).
While this result is obviously quite preliminary, it was a good demonstration that
it is not easy distinguishing machine from human errors, even for technically sophisticated observers.

\section{Conclusions}

We have discussed methodological issues and reported first findings when comparing automatic conversational speech transcriptions
to human performance, using data generated by our recent efforts to reach human parity in CTS recognition.
While an exact characterization of the human benchmark remains a moving target that is subject to debate, our results so far
have shown that machine transcription errors track those made by humans in several important aspects.
At the speaker (as well as corpus) level the two error rates are strongly correlated, suggesting that common underlying factors
in the speech data determine transcription difficulty for both humans and ASR systems.
(A detailed characterization of those factors has precedent in ASR research and should be revisited while also considering
human performance.)
A partial overlap of Switchboard training and test speakers seems to have no major effect on error rates.
We also find that the most frequent error patterns involve the same short function words and discourse particles for both humans
and machines.  The one notable exception is that ASR tends to confuse filled pauses and backchannels,
a functional distinction that humans need to be very good at pragmatically.
An informal Turing-like test also demonstrated that error patterns in the two types of transcriptions are not obviously distinguishable.
Overall, we conclude that recent advances in ASR technology have not only achieved remarkable levels of accuracy,
but also generate results that are qualitatively surprisingly similar to professional human transcriber output.

\footnotesize

\section{Acknowledgments}

We thank our coauthors and collaborators on the Human Parity project: X. Huang, F. Seide, M. Seltzer, W. Xiong, D. Yu, and G. Zweig.
Thanks to K. Riedhammer for sharing metadata on train/test speaker overlap.

\normalsize

\bibliographystyle{ieee-shortnames}

\bibliography{strings,refs}

\end{document}